\documentclass[10pt,twocolumn,letterpaper]{article}

\usepackage[pagenumbers]{iccv} 

\definecolor{iccvblue}{rgb}{0.21,0.49,0.74}
\usepackage[pagebackref,breaklinks,colorlinks,allcolors=iccvblue]{hyperref}
\usepackage{array}
\usepackage{tabularx}
\newcolumntype{Y}{>{\centering\arraybackslash}X}
\usepackage{colortbl}
\usepackage{graphicx}
\usepackage{wrapfig}
\definecolor{yellow}{rgb}{1, 1, 0.7}
\definecolor{orange}{rgb}{1, 0.85, 0.7}
\definecolor{red}{rgb}{1, 0.7, 0.7}
\definecolor{normalred}{rgb}{1, 0, 0}
\usepackage{enumitem}
\usepackage[ruled]{algorithm2e} 
\usepackage{xcolor}

\def\confName{ICCV}
\def\confYear{2025}

\title{GAP: Gaussianize Any Point Clouds with Text Guidance}

\author{
Weiqi Zhang$^{1*}$, Junsheng Zhou$^{1*\dag}$, Haotian Geng$^{1*}$, Wenyuan Zhang$^{1}$, Yu-Shen Liu$^{1\dag}$\\
School of Software, Tsinghua University, Beijing, China$^1$\\
{\tt\small \{zwq23, zhou-js24, genght24, zhangwen21\}@mails.tsinghua.edu.cn}\\
{\tt\small liuyushen@tsinghua.edu.cn}\\
}

\begin{document}

\twocolumn[{%
\renewcommand\twocolumn[1][]{#1}%
\maketitle
\begin{center}
\centering
\captionsetup{type=figure}
\vspace{-10mm}
\includegraphics[width=\linewidth]{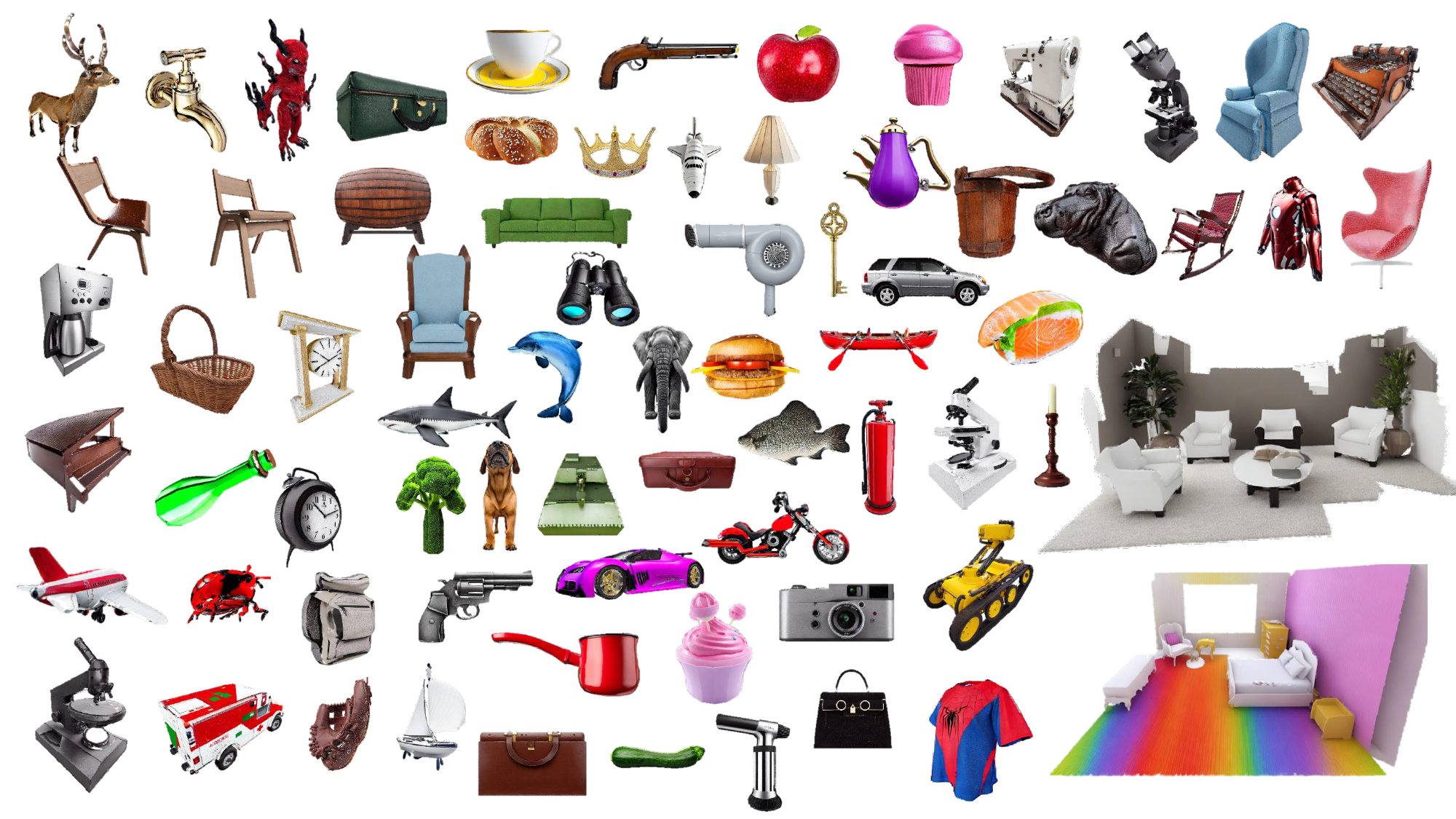}
\vspace{-8mm}
\captionof{figure}{
GAP gaussianizes point clouds into high-fidelity 3D Gaussians with diverse appearances. Left: Examples of text-guided Gaussian generation from object-level point cloud. Bottom-right: Scene-level results with prompts 'A modern lounge' and 'A rainbow bedroom'.
}
\vspace{-3mm}
\label{fig:teaser}
\end{center}%
}]

\if TT\insert\footins{\footnotesize{
*Equal contribution. $\dag$ Corresponding authors. This work was supported by National Key R\&D Program of China (2022YFC3800600), and the National Natural Science Foundation of China (62272263), and in part by Tsinghua-Kuaishou Institute of Future Media Data. Junsheng Zhou is also partially funded by Baidu Scholarship.}}\fi

\begin{abstract}
3D Gaussian Splatting (3DGS) has demonstrated its advantages in achieving fast and high-quality rendering. As point clouds serve as a widely-used and easily accessible form of 3D representation, bridging the gap between point clouds and Gaussians becomes increasingly important. Recent studies have explored how to convert the colored points into Gaussians, but directly generating Gaussians from colorless 3D point clouds remains an unsolved challenge. In this paper, we propose GAP, a novel approach that gaussianizes raw point clouds into high-fidelity 3D Gaussians with text guidance. Our key idea is to design a multi-view optimization framework that leverages a depth-aware image diffusion model to synthesize consistent appearances across different viewpoints. To ensure geometric accuracy, we introduce a surface-anchoring mechanism that effectively constrains Gaussians to lie on the surfaces of 3D shapes during optimization. Furthermore, GAP incorporates a diffuse-based inpainting strategy that specifically targets at completing hard-to-observe regions. We evaluate GAP on the Point-to-Gaussian generation task across varying complexity levels, from synthetic point clouds to challenging real-world scans, and even large-scale scenes. Project Page: \url{https://weiqi-zhang.github.io/GAP}.
\end{abstract}    
\vspace{-0.6cm}
\section{Introduction}
\label{sec:intro}
Point clouds serve as a fundamental representation in 3D computer vision, playing a crucial role across various domains~\cite{takeshi2024multipull,zhou2023uni3d,li2024LDI,Zhou2023VP2P,3DAttriFlow}, e.g., autonomous driving, augmented/virtual reality and robotics. With recent advances in 3D scanning devices, such as LiDAR sensors and depth cameras, point clouds have bridged the gap between the physical and digital worlds. However, it still remains a research challenge to effectively transform the geometries of raw point clouds into high-quality 3D appearances that maintain structural fidelity while providing visual-appealing renderings.
 
For high-quality 3D visualization, mesh-based representation has long been the standard approach. However, such a representation faces two major limitations: (1) for meshes with dense faces, the constrained texture resolution limits the final rendering quality, and (2) the heavy reliance on UV unwrapping \cite{zhou2006mesh} introduces additional complications such as texture overlapping, fragmentation, and distortion issues. While these limitations can be addressed with careful manual intervention, they present significant obstacles in fully automated pipelines. Recent advances in 3D Gaussian Splatting (3DGS) \cite{kerbl20233dgs} have revolutionized neural rendering by offering an efficient and high-quality alternative to NeRF-based \cite{mildenhall2020nerf} or mesh-based representations. Moreover, 3DGS eliminates the need for explicit UV parameterization, which makes it particularly attractive for real-world applications. 

Althought several attempts have been made to bridge point clouds and 3DGS, existing approaches still face several significant limitations. For example, Large Point-to-Gaussian \cite{lu2024large} model trains a feedforward network for Gaussian primitive generation, but it requires point cloud inputs with color attributes. DiffGS \cite{zhou2024diffgs} approaches this challenge by learning a reconstruction scheme from points to Gaussians, yet struggles in generalizing to generate diverse and high-quality 3D appearances.

To address these challenges, we propose GAP, a novel approach that generates high-quality Gaussian primitives by Gaussianizing 3D raw point clouds. GAP leverages both geometric information from input point clouds and appearance guidance from pretrained text-to-image diffusion models. Specifically, we first introduce a progressive generation scheme that optimizes Gaussian primitives across multiple viewpoints by leveraging a depth-aware text-to-image diffusion model. To ensure geometric accuracy, we design a surface-anchoring mechanism that effectively constrains Gaussians to lie on object surfaces during optimization, leading to Gaussian generations consistent to the geometry. After optimization, the generated high-quality Gaussians can cover most of the surface, however, there are still some unseen areas that require further processing. To address this, we propose a diffuse-based Gaussian inpainting strategy that gaussianizes the unseen points by leveraging the spatial relationships and geometric consistency of the visible Gaussians. To this end, GAP generates high-fidelity 3D Gaussians that maintain both geometric accuracy and visual quality.

We evaluate GAP extensively across diverse datasets, including both synthetic and real-world scanned point clouds of objects and scenes. Comprehensive experiments demonstrate that our method consistently outperforms state-of-the-art alternatives in terms of visual quality. We believe GAP opens new possibilities for Point-to-Gaussian generation, bridging the gap between widely-used, easily accessible point cloud data and high-quality 3D Gaussian representations. Our contributions can be summarized as follows:

\begin{itemize}
\item We proposed GAP, a novel framework that gaussianizes raw point clouds into high-quality Gaussian primitives. GAP introduces both geometric priors and text guidance with large text-to-image diffusion models to generate diverse and visual-appealing appearances from point clouds.   
\item We design a Gaussian optimization framework that progressively optimizes Gaussian attributes across multiple viewpoints, with a surface anchoring constraint to ensure geometric accuracy. A diffuse-based Gaussian inpainting strategy is further introduced to handle occluded regions.
\item Comprehensive evaluations under synthetic and real-scanned point cloud datasets of objects and scenes demonstrate that GAP significantly outperforms the state-of-the-art methods.
\end{itemize}

\vspace{-0.2cm}
\section{Related Work}
\label{sec:formatting}
\vspace{-0.1cm}

\subsection{Texture Generation} 
\vspace{-0.2cm}
The advent of deep learning has revolutionized texture generation for 3D models. Early learning-based approaches primarily utilized GANs \cite{goodfellow2020generative,mirza2014conditional,radford2015unsupervised,zhou20223d}, while recent methods \cite{yu2023texture,cao2023texfusion,xiang2024make, liu2024text, jiang2024flexitex} leverage large-scale text-to-image diffusion models \cite{ho2020denoising,rombach2022high} as powerful priors for high-fidelity texture synthesis. A series of works \cite{metzer2023latent, chen2023fantasia3d,yeh2024texturedreamer} adopts Score Distillation Sampling \cite{poole2022dreamfusion} as their optimization strategy for texture generation, iteratively refining textures through optimizing rendered images with respect to text prompts. Another stream of research \cite{richardson2023texture,chen2023text2tex,tang2024intex} proposes efficient texture synthesis through depth-guided inpainting, where textures are progressively generated along specified viewpoints. Additionally, some approaches \cite{zeng2024paint3d,bensadoun2024meta, cheng2024mvpaint} focus on multi-view generation with geometric guidance, followed by UV-space refinement. However, maintaining texture continuity across UV seams remains challenging due to the discontinuous nature of UV mapping. Despite these advances, UV distortion and cross-view consistency remain challenging, particularly for complex objects.

\subsection{Rendering-Driven 3D Representation} 
\vspace{-0.2cm}
While mesh-based representations \cite{botsch2010polygon, seng2009realistic} remain the standard for 3D visualization, they face limitations in texture resolution and UV parameterization \cite{levy2023least,  sander2002signal}. Remarkable progress has been achieved in the field of novel view synthesis with the proposal of Neural Radiance Fields (NeRF) \cite{mildenhall2020nerf}. Through volume rendering \cite{drebin1988volume} optimization, NeRF achieves outstanding view synthesis quality, though its computational overhead during rendering is considerable. 3D Gaussian Splatting (3DGS) has emerged as an advanced 3D representation which shows convincing performance in real-time rendering \cite{kerbl20233dgs,tang2023dreamgaussian, li2024dngaussian, yi2023gaussiandreamer, yu2024mip, zhang2024learning,huang2024neusurf,zhang2023fast}. By representing scenes with a set of 3D Gaussian primitives, 3DGS achieves both high-quality rendering and efficient real-time performance. 

\subsection{3DGS Generation Methods} 
\vspace{-0.2cm}
A series of studies have explored this direction by integrating neural implicit representations\cite{noda2025learning,udiff,zhou2024cappami,li2023NeAF,zhou2023levelset,fastn2n,BaoruiTowards,sparserecon,Sparis}, laying the groundwork for more expressive and efficient 3D modeling. Building on this progress, the advancement of 3D Gaussian Splatting has further accelerated interest in developing effective generative models for 3DGS, making it a rapidly emerging research focus. A series of studies \cite{xu2024grm,zou2023triplane,hong2023lrm,zhang2024gs, zhang2025nerfprior, zhang2025monoinstance, li2025gaussianudf, zhang2024gspull,han2024binocular,zhou2024DeepPriorAssembly} have explored image-based reconstruction without generative modeling, which fundamentally limits their ability to generate diverse shapes. These methods also lack point-conditioned generation capabilities. Recent point cloud-to-Gaussian conversion approaches \cite{lu2024large} rely heavily on the availability of RGB point clouds as input. While Gaussian painter \cite{zhou2024gaussianpainter} uses reference images for stylization, it lacks precise control over the final appearance. This highlights the need for a framework generating high-quality Gaussians from point clouds with flexible appearance control.
\vspace{-0.2cm}
\section{Method}
\label{sec.3}

\begin{figure*}[!t]
  \centering
  \includegraphics[width=\textwidth]{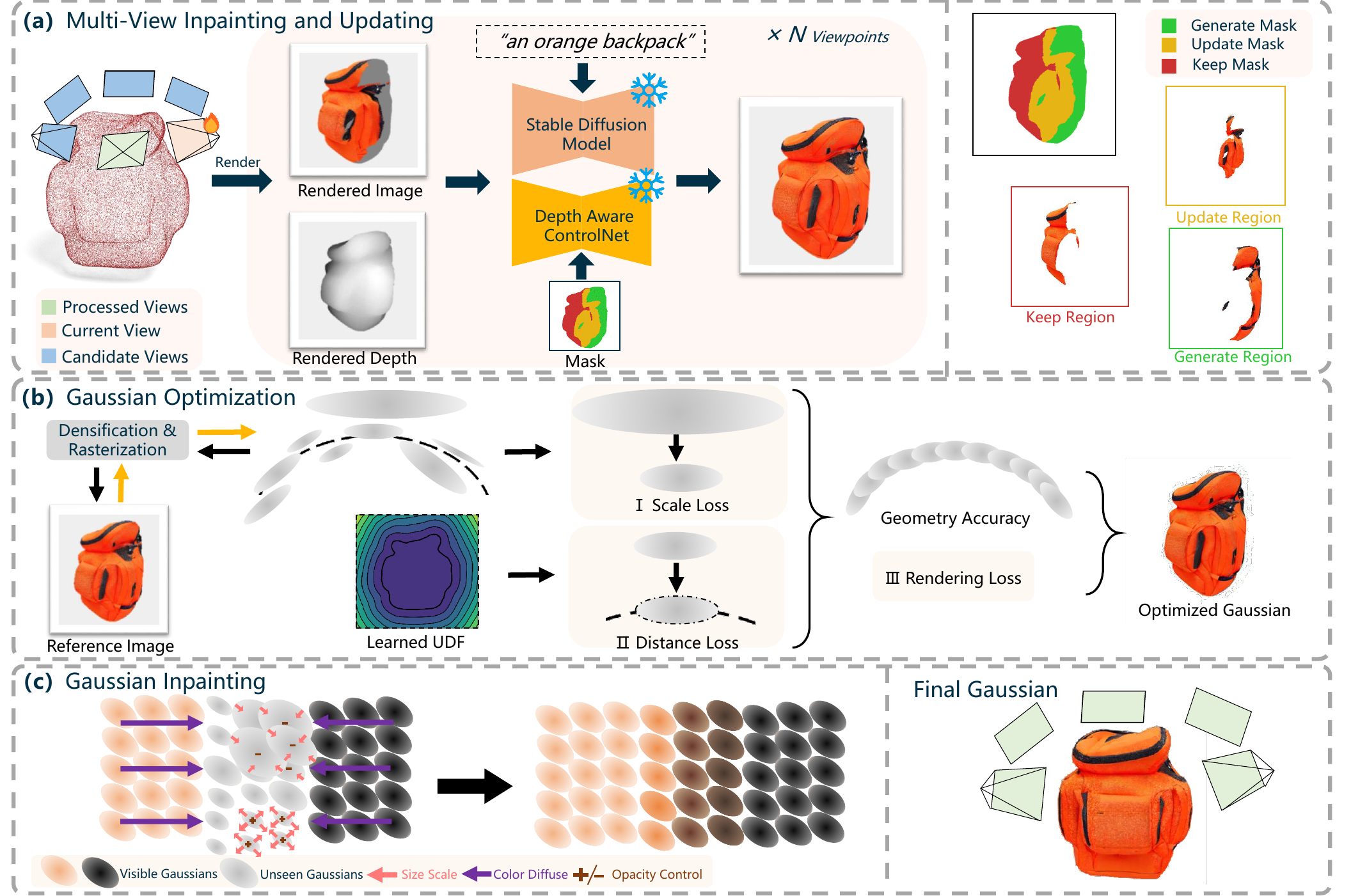}
    \vspace{-0.8cm}
  \caption{\textbf{Overview of GAP. (a)} We rasterize the Gaussians though an unprocessed view, where a depth-aware image diffusion model is used to generate consistent appearances using the rendered depth and mask with text guidance. The mask is dynamically classified as generate, keep, or update based on viewing conditions. \textbf{(b) } The Gaussian optimization includes three constraints: the Distance Loss and Scale Loss introduced to ensure geometric accuracy, and the Rendering Loss that ensures high-quality appearance. \textbf{(c) } The Gaussian inpainting strategy which diffuses the geometric and appearance information from visible regions to hard-to-observe areas, considering local density, spatial proximity and normal consistency.}
  \vspace{-0.4cm}
  \label{fig:overview}
\end{figure*}

We introduced GAP, a novel method that establishes a bridge between raw point clouds and 3D Gaussians by neural gaussianizing. Given an input point cloud $P = \{p_i\}_{i=1}^N$, our goal is to generate Gaussians $G = \{g_i\}_{i=1}^M$ from $P$, conditioned on the text prompt $c$. The overview of GAP is shown in Fig.~\ref{fig:overview}. We begin by previewing Gaussian Splatting, along with the initialization strategy in Sec.~\ref{sec.3.1}. In Sec.~\ref{sec.3.2}, we present a progressive Gaussian generation scheme that utilizes a powerful text-to-image diffusion model to generate or inpaint images from a given viewpoint. We further introduce a Gaussian optimization strategy which learns Gaussian attributes from the generated images representing high-fidelity appearance, in Sec. \ref{sec.3.3}. While the object is largely observable from various viewpoints, certain regions remain difficult to capture. To address this, we introduce a diffuse-based Gaussian inpainting method in Sec.~\ref{sec.3.4}.

\subsection{Gaussian Initialization}
\label{sec.3.1}
\noindent\textbf{Preview 3D Gaussian Splatting.} 3D Gaussian Splatting (3DGS)~\cite{kerbl20233dgs} is a modern representation technique that models 3D shapes or scenes through a collection of Gaussian primitives. Each Gaussian $g_i$ is defined by a set of parameters that characterize its geometry and appearance properties. The geometry of $g_i$ is mathematically defined by its center position $\sigma_i \in \mathbb{R}^3$ and a covariance matrix $\Sigma_i$, formulated as:
\begin{equation}
    g_i(x) = \exp\left(-\frac{1}{2} (x - \sigma_i)^T \Sigma^{-1} (x - \sigma_i)\right).
\end{equation}

The covariance matrix $\Sigma_i$ is constructed from a rotation matrix $r_i \in \mathbb{R}^4$ and a scale matrix $s_i \in \mathbb{R}^3$ ($\Sigma_i = r_i s_i s_i^T r_i^T$). $\Sigma_i$ determines the Gaussian's shape, orientation, and range in space.  Beyond geometry, each Gaussian encompasses visual attributes including an opacity term $o_i$ and view-dependent color properties $c_i$, implemented as spherical harmonics.

\noindent\textbf{Initialization Scheme.} When generating Gaussians from an input point cloud $P = \{p_i\}_{i=1}^N$, we initialize the center positions $\sigma_i$ of Gaussian primitives as the spatial coordinates of the points. This direct spatial mapping provides fine initial geometries for Gaussians, which roughly represent the underlying 3D surfaces. To better exploit the inherent geometric information embedded in the point cloud, we employ CAP-UDF to learn a neural Unsigned Distance Field (UDF) \cite{chibane2020neural} $f_u$ from the point cloud and derive point normals $N = \{n_i\}_{i=1}^N$ through gradient inference:
\begin{equation}
    n_i = \frac{\nabla f_u(p_i)}{\Vert \nabla f_u(p_i) \Vert}. 
\end{equation}


 Instead of vanilla 3DGS, we adopt 2D Gaussian Splatting (2DGS) \cite{huang20242d} as our representation. The key idea of 2DGS is to replace 3D Gaussian ellipsoids with 2D-oriented Gaussian disks for scene representation, demonstrating better performances in representing detailed local geometries. 2DGS inherently encodes the normal as the disk orientation. We initialize the rotation matrix $r_i$ of each Gaussian using its normal $n_i$ from the field $f_u$, ensuring that each 2D Gaussian disk is accurately aligned to the correct orientation, providing a good initialization for subsequent optimization. 

\subsection{Multi-View Inpainting and Updating}
\label{sec.3.2}
For a sequence of specified viewpoints $\{v_j\}_{j=1}^K$, we progressively generate the visual appearance at each perspective to optimize the associated Gaussians. Using the learned UDF field, we employ ray marching techniques to compute the depth value for each pixel on the depth map $D_j$.  As shown in Fig.~\ref{fig:overview}(a), we render an image $I_j$ from a specific viewpoint $v_j$. The rendered image $I_j$, along with its corresponding depth map $D_j$, mask $M_j$ and text prompt $c$, are fed into the depth-aware inpainting model. 


\noindent\textbf{Depth-aware Inpainting Model.} 
We leverage a depth-aware inpainting diffusion model \cite{rombach2022high, zhang2023controlnet} as the appearance generation model. By integrating depth information into the diffusion-based inpainting process, the model enables more geometrically consistent image generation. Its encoder $\mathbb{E}$ operates by first encoding the masked image $I$ concatenated with the depth map $D$ into a latent code $z_0$. The initial encoding is:
\begin{equation}
    z_0 = \mathbb{E}\left[I \parallel D \right].
\end{equation}

The process gradually degrades the initial latent code through a series of noise-adding operations. At each timestep $t$, the model add Gaussian noise according to a variance schedule defined by $\beta_t$. The transformation follows a probabilistic distribution:
\begin{equation}
    z_t \mid y, \, g_{\phi}\left(y, t, I \parallel D\right) \sim \mathcal{N}\left(\sqrt{1 - \beta_t}\, z_{t-1},\; \beta_t\, \mathbf{I}\right), 
\end{equation}
where $y$ is text embeddings, and $g_{\phi}$ is ControlNet function processing the image-depth input. 


To maintain generation consistency, mask blending is operated at each timestep. Specifically, the latent encoding $z_t$ at timestep $t$ is combined with the masked region encoding $z_{M, t}$ according to masks $M$. The mask blending operation ensures that the content in the unmasked regions is well preserved. It can be formulated as:
\begin{equation}
    z_t \leftarrow z_t \odot M + z_{M, t} \odot \left(1 - M\right).
\end{equation}

\noindent\textbf{Updating Scheme for Inpainting.}
For the same area of the 3D shape, the inpainting model may generate varying appearances. We implemented an updating scheme that allows us to refine previously processed regions when more favorable viewing angles become available. Hence, masks $M$ are divided into three distinct regions based on their visibility from the current viewpoint $v_j$: generate mask $M_{generate}$, keep mask $M_{keep}$ and update mask $M_{update}$. 

The generate masks $M_{generate}$ refer to blank areas that have never been generated before. The keep masks $M_{keep}$ are those that have been processed before and the current viewpoint does not provide better viewing conditions. The calculation of the update mask $M_{update}$ involves evaluating whether to refresh a region based on the similarity between its viewing directions and normals. Specifically, we define a similarity mask $M_{similarity}$ to quantify the observability of surface details from different viewing angles. For a viewpoint $v_j$, the similarity mask value is computed as the cosine similarity between the viewing direction $d_j$ and the point normal $N$: $M_{similarity} = d_j \cdot N$. A region should be updated when the current view provides a better observation angle than any other view:
\begin{equation}
    M_{update}^j = \begin{cases} 1, & \text{if } M_{similarity}^j > M_{similarity}^{others} \\ 0, & \text{otherwise}. \end{cases} \vphantom{\int}
\end{equation}
The final inpainting $I_{inpaint}$ is generated by combining two different denoising processes: a stronger denoising for newly generated regions (generate masks) and a weaker denoising for regions requiring updates (update masks). The final appearance is achieved as:
\begin{equation}
    I \leftarrow I_{inpaint} \odot (1 - M_{keep}) + I \odot M_{keep}.
\end{equation}

\subsection{Gaussian Optimization}
\label{sec.3.3}
\begin{figure}[!t]
    \centering
    \includegraphics[width= \linewidth, alt={A detailed description of the method illustrated in the image}]{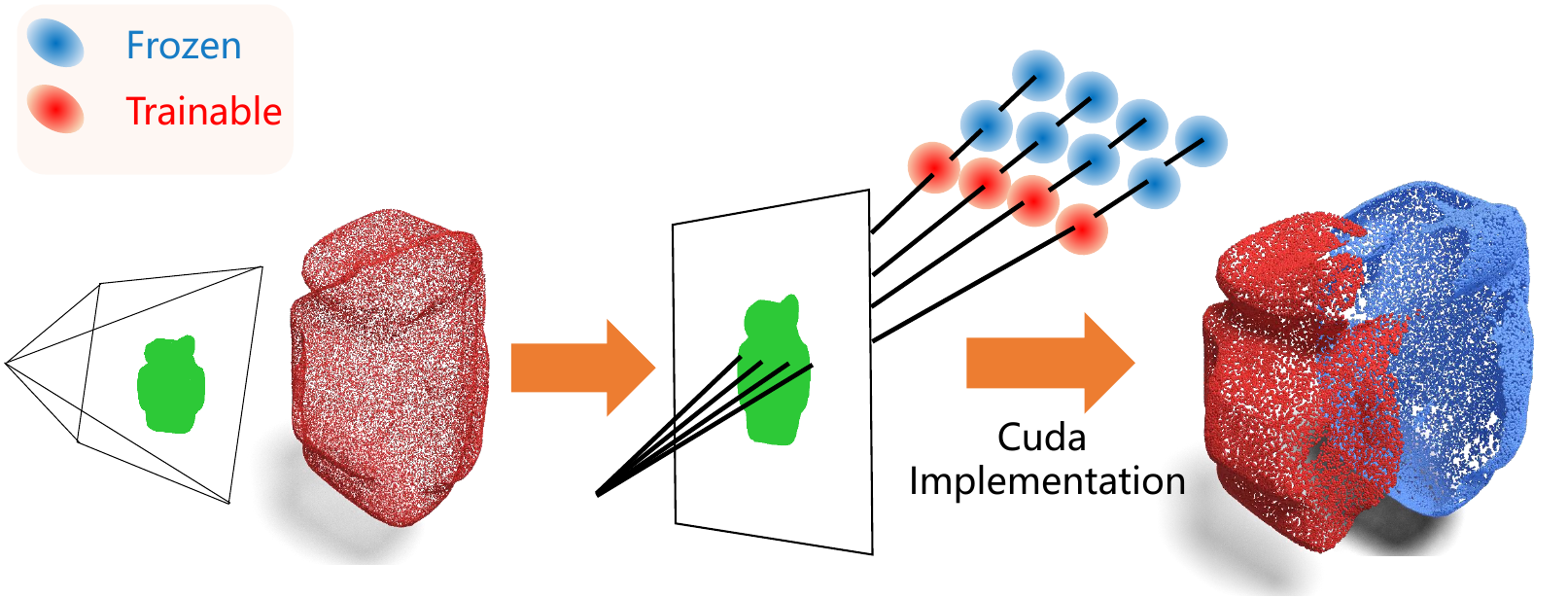}
    \vspace{-0.8cm}
    \caption{\textbf{Gaussian Selection scheme.} We identifies the first intersecting Gaussian along each viewing ray within generate or update masks, implemented with CUDA for efficient processing.}
    \label{fig:selection}
    \vspace{-0.6cm}
\end{figure}

For a given viewpoint $v_j$, we can now generate the appearance $I_j$ with the powerful inpainting model. The Gaussians $G$ can be optimized through $I_j$. Unlike the vanilla 3DGS fitting scheme that optimizes Gaussian attributes through multiple iterations across different viewpoints, GAP operates only a single optimization pass per viewpoint, which leads to more robust Gaussian generations faithfully representing the high-quality appearance $I_j$. Specifically, in each view-specific optimization step, we focus exclusively on optimizing the Gaussians that represent the nearest visible surface layer from the current viewpoint, without modifying the Gaussians on the back-facing surfaces, as shown in Fig.~\ref{fig:selection}. To this end, we implement a Gaussian selection scheme that identifies the first intersecting Gaussian along each viewing ray originating from pixels within the generate or update mask. To manage the computational intensity of processing numerous rays, we develop a CUDA \cite{cuda} implementation that exploits GPU parallelism. accelerating the Gaussian selection process to just 3 seconds.

\noindent\textbf{Surface-anchoring Mechanism.} During Gaussian optimization, Gaussians that float away from their expected surface positions introduce significant challenges for multi-view inpainting and updating. These Gaussians produce incorrect occlusion relationships in subsequent viewpoints, resulting in distorted masks and further degrading the quality of generation and inpainting. To this end, we introduce a surface-anchoring mechanism in terms of a distance loss which aligns Gaussians with the zero-level set of the learned unsigned distance field. Practically, we constrain distance value at each Gaussian center, queried from $f_u$, to be close to zero during optimization. The distance loss is formulated as: 
\begin{equation}
    \mathcal{L}_{\mathrm{Distance}} = \| f_u(\sigma_i)\|_2.
\end{equation}

\noindent\textbf{Scale Constraint.} During optimization from a single viewpoint, some oversized Gaussians may lead to incorrect geometries which adversely affect the inpainting results of subsequent views. To address this issue, we introduce a scale loss that constrains the maximum value of $s_i$ for each Gaussian. The \textit{Scale Loss} is defined as:\
\begin{equation}
     \mathcal{L}_{\mathrm{Scale}} = \left( \min(\max(s_i), \tau) - \max(s_i) \right)^2,
\end{equation}
where $\tau$ is a predefined threshold value. The scale loss effectively prevents Gaussians from growing excessively large while still allowing sufficient flexibility to model the appearance. 

\noindent\textbf{Rendering Constraint.} Following 3DGS \cite{kerbl20233dgs}, we also employ the \textit{Rendering Loss} during optimization. 
The rendering constraint consists of an $L1$ loss term and a D-SSIM term with weights of 0.8 and 0.2 respectively:
\begin{equation}
    \mathcal{L}_{\mathrm{Rendering}} = 0.8 L_1(I_j^{\prime},I_j) + 0.2 L_{D-SSIM}(I_j^{\prime},I_j),
\end{equation}
where $I_j^{\prime}$ is the rendered image. With the balanced weight $\alpha$ and $\beta$, the final optimization objective can be formulated as:
\begin{equation}
    \mathcal{L} = \mathcal{L}_{\mathrm{Rendering}} + \alpha\mathcal{L}_{\mathrm{Distance}} + \beta\mathcal{L}_{\mathrm{Scale}}.
\end{equation}

\subsection{Diffuse-based Gaussian Inpainting}
\label{sec.3.4}

\begin{wrapfigure}{r}{0.5\linewidth}
    \centering
    \resizebox{1\linewidth}{!}{
    \hspace{-0.3cm}
    \includegraphics[width=\linewidth]{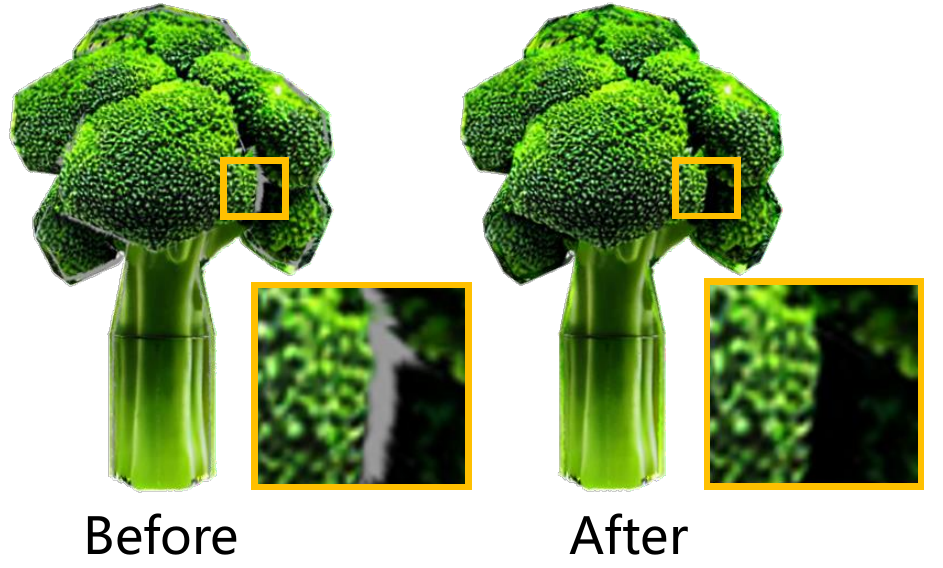}}
    \vspace{-0.8cm}
    \caption{The Gaussian inpainting approach effectively completes the unseen regions by propagating properties from visible Gaussians.}
    \label{fig:woinpainting}
    \vspace{-0.4cm}
\end{wrapfigure}

Even with comprehensive multi-view capturing from densely sampled viewpoints, certain regions of the 3D object are still challenging to observe. As shown in Fig.~\ref{fig:overview}(c), to model the appearances of the unseen areas, we propose a diffuse-based Gaussian inpainting approach. 
Our method effectively recovers missing appearance in the final representation, as shown in Fig.~\ref{fig:woinpainting}. 
Our approach operates inpainting directly in 3D space, leveraging the inherent structure and spatial relationships of the visible Gaussians. Using the Gaussian selection scheme across multiple viewpoints, we can effectively identify the unseen Gaussians $G^{\prime} = \{g_j^{\prime}\}_{j=1}^{M^{\prime}}$, which are not optimized at any view. For the unseen Gaussians, whose positions and normal directions have already been well initialized through the Gaussian initialization scheme proposed in Sec.~\ref{sec.3.1}, we primarily focus on predicting their remaining properties, such as color, scale, and opacity.

\begin{figure*}[!h]
    \centering
    \vspace{-1.2cm}
    
    \includegraphics[width= \linewidth, alt={A detailed description of the method illustrated in the image}]{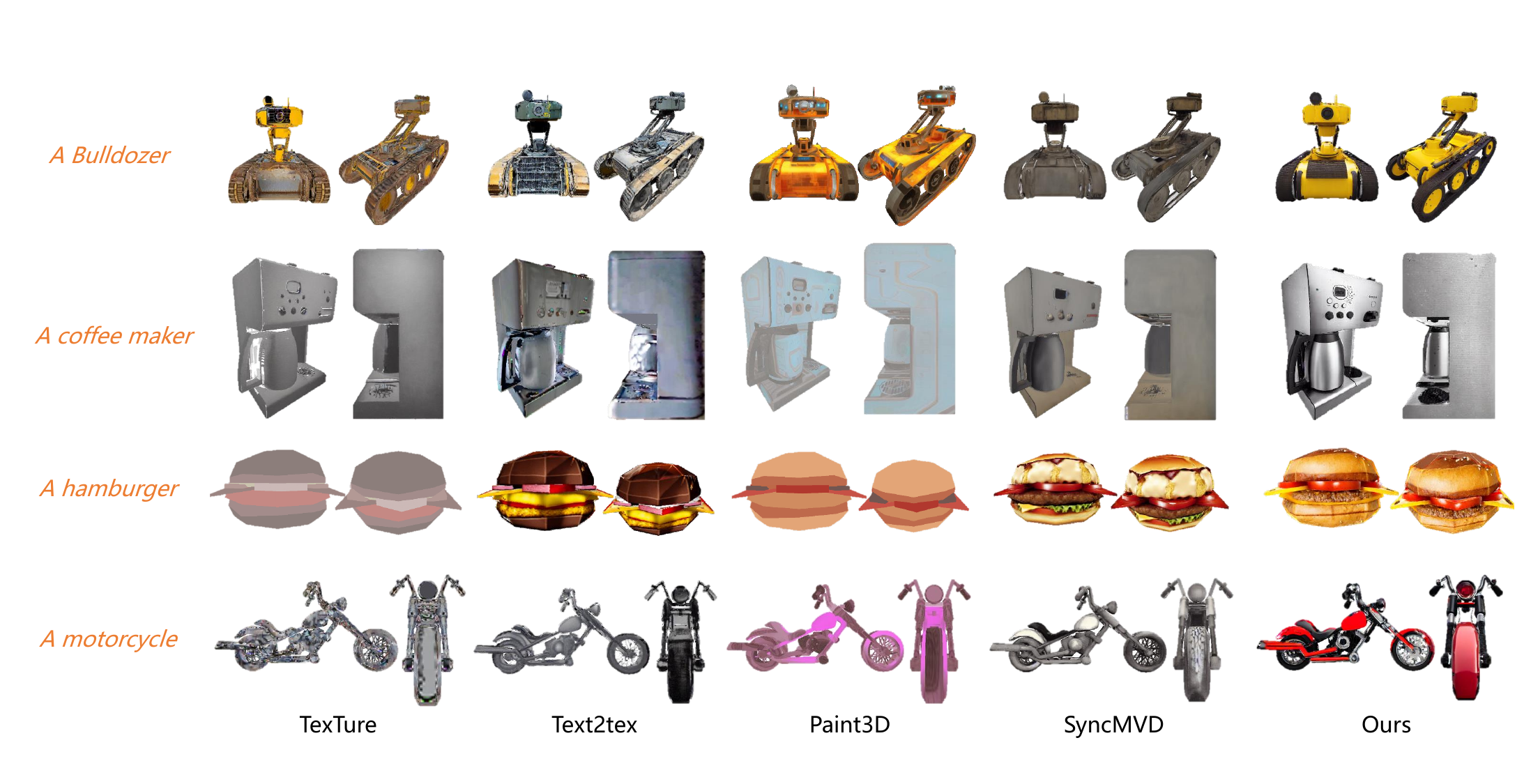}
    \vspace{-1.2cm}
    \caption{Visual comparison of text-guided appearance generation results on the Objaverse dataset. }
    \label{fig:quality_objaverse}
    \vspace{-0.5cm}
\end{figure*}


\noindent\textbf{Color Diffuse.} To predict the color attributes of the unseen regions, we implement a diffusion mechanism that propagates the attributes of nearby Gaussians. For each unseen Gaussian $g_j^{\prime}$,  we first locate its $L$ nearest optimized neighbor Gaussians as the reference. We design a weighting strategy that incorporates spatial proximity, geometric consistency, and opacity reliability during color diffuse. Let $o_{max}$ be the maximum opacity value among all neighbor Gaussians. For each valid neighbor $g_i$ of the unseen Gaussian $g_j^{\prime}$, we define its color weight $\lambda_i$ as follows: when the angle between the normals of $g_i$ and $g_j^{\prime}$ is less than 60 degrees, i.e., $(\mathbf{n}_i \cdot \mathbf{n}_j)>0.5$, the weight is calculated as:
\begin{equation}
    \lambda_i = \frac{1/d_i}{\sum_{k=1}^{L}1/d_k} \cdot (\mathbf{n}_i \cdot \mathbf{n}_j) \cdot \frac{o_i}{o_{max}}. \\
\end{equation}

Otherwise, the weight is set to 0. The distance term $1/d_i$ prevents the far Gaussians with inconsistent appearances to largely affect the color, while the normal consistency term $(\mathbf{n}_i \cdot \mathbf{n}_j)$ preserves geometric features by prioritizing color propagation between Gaussians with similar surface orientations. The opacity reliability term $o_i/o_{max}$ ensures that Gaussians with higher opacity values have a stronger influence on the color prediction. Finally, the color $c_j^{\prime}$ of the unseen Gaussian $g_j^{\prime}$ can be formulated as:
\begin{equation}
    c_j^{\prime} = \frac{\sum_{i=1}^{L}(c_i * \lambda_i)}{\sum_{i=1}^{L}\lambda_i}.
\end{equation}

\noindent\textbf{Size Scale.}
To predict appropriate scales for the unseen Gaussians $g_j^{\prime}$, we consider the $L$ nearest neighbors (including both optimized and unseen Gaussians). The scale is adjusted based on the spatial proximity of these neighbors. The scale $s_j^{\prime}$ of an unseen Gaussian  is computed as:
\begin{equation}
    s_j^{\prime} = \log(\frac{\sum_{i=1}^{L}d_i}{L}),
\end{equation}
where $d_i$ represents the distance between the unseen Gaussian $g_j^{\prime}$ and its neighbor $g_i$. We incorporate distance weighting, as larger distances indicate sparser regions that require larger scales.

\noindent\textbf{Opacity Control.}
For predicting the opacity $o_j^{\prime}$ of an unseen Gaussian $g_j^{\prime}$, we employ a density-based control mechanism. The opacity within a radius $\rho$ is inversely proportional to the local Gaussian density. The opacity $o_j^{\prime}$ of an unseen Gaussian $g_j^{\prime}$ is computed as: 
\begin{equation}
    o_j^{\prime} = \frac{o_0}{max(1, P/P_0)},
\end{equation}
where $o_0$ is a base opacity value, $P$ is the number of neighboring Gaussians within a specified radius $\rho$, and $P_0$ is a reference density threshold. The opacity control scheme ensures that regions with higher Gaussian density have lower opacity values, preventing over-accumulation of color while maintaining proper surface coverage.

\begin{figure*}[]
    \centering
    \includegraphics[width=\linewidth, alt={A detailed description of the method illustrated in the image}]{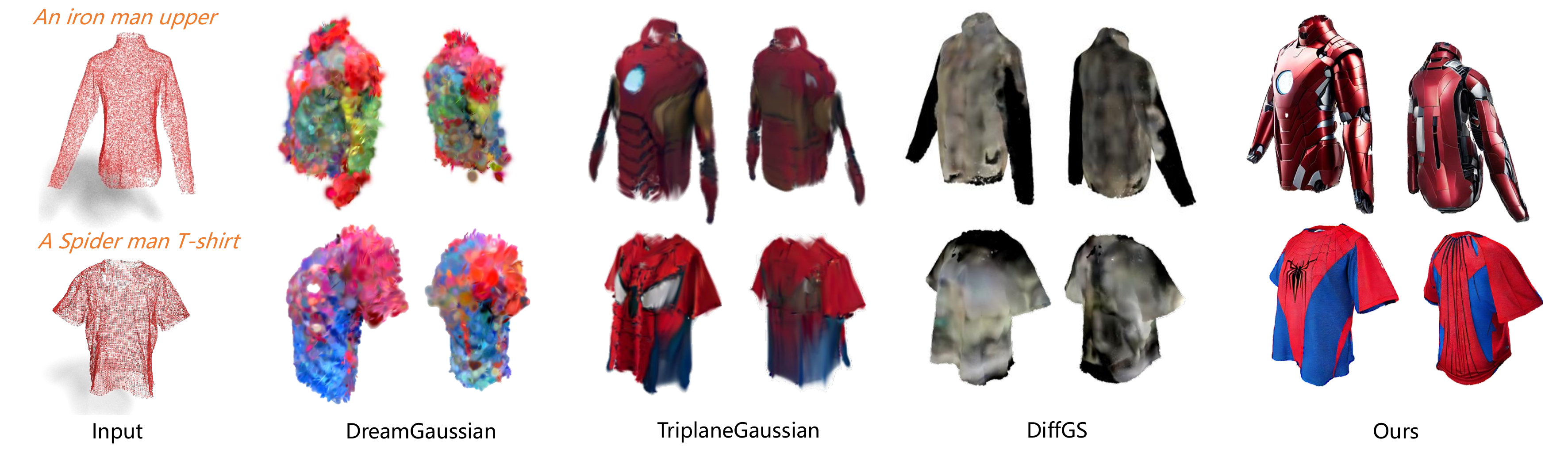}
    \vspace{-0.8cm}
    \caption{Visual comparison of point-to-Gaussian generation results on DeepFashion3D.}
    \label{fig:point2gaussian}
    \vspace{-0.4cm}
    
\end{figure*}

\section{Experiments}
\label{sec.4}

We first evaluate GAP's core capability of text-driven appearance generation in Sec.~\ref{sec.4.1}. In Sec.~\ref{sec.4.2}, we compare GAP's performance specifically on the Point-to-Gaussian generation task with other Gaussian generation methods. Next, we further validate GAP's capability on real-world scanned point clouds, where the inputs are often incomplete in Sec.~\ref{sec.4.3}. In Sec.~\ref{sec.4.4}, we showcase GAP's scalability by applying it to scene-level point clouds. Finally, the ablation studies are shown in Sec.~\ref{sec.4.5}. 

\subsection{Text-Driven Appearance Generation}
\label{sec.4.1}
\noindent\textbf{Datasets and Metrics.} Following prior works \cite{chen2023text2tex, richardson2023texture}, we conduct experiments on the curated subset of the Objaverse \cite{objaverse} dataset containing 410 textured meshes across 225 categories. Unlike previous methods that require perfect meshes, we only use a sampled point cloud of $100K$ points as input. 
We employ three complementary metrics: Fréchet Inception Distance (FID) \cite{yu2021frechet} and Kernel Inception Distance (KID $\times 10^{-3}$) \cite{binkowski2018demystifying} for assessing image quality, and CLIP Score \cite{radford2021learning} for measuring text-image alignment. All methods use identical text prompts describing each object.
We render all objects at a high resolution of $1024\times1024$ pixels from fixed viewpoints.

\noindent\textbf{Baselines.} 
For visual appearance, we compare GAP with state-of-the-art 3D texture generation methods: TexTure \cite{richardson2023texture}, Text2Tex \cite{chen2023text2tex}, Paint3D \cite{zeng2024paint3d}, and SyncMVD \cite{liu2024text}, all of which rely on UV-mapped meshes. And the original meshes in the subset of the Objaverse dataset include artist-created UV maps. For a fair comparison with those methods under the same conditions of point cloud inputs, we reconstruct meshes from the input point clouds using the traditional Ball-Pivoting Algorithm (BPA) \cite{bernardini1999ball} and SOTA learning-based method CAP-UDF \cite{Zhou2022CAP-UDF}. We then generate UV maps through xatlas \cite{xatlas} unwrapping. 



\noindent\textbf{Comparison.} The quantitative comparison in Tab.~\ref{tab:quantitative_objaverse} shows that GAP outperforms previous state-of-the-art methods. Unlike approaches relying on artist-created UV maps, GAP leverages Gaussian Splatting for inherently higher rendering quality. The performance gap is even more pronounced compared to baselines using reconstructed meshes, which suffer from topological ambiguities, connectivity errors, and geometric distortions. These issues, compounded by dense mesh reconstructions and automated UV unwrapping, often result in severe texture artifacts. In contrast, GAP bypasses UV parameterization by directly optimizing Gaussian primitives in 3D space.
As shown in Fig.~\ref{fig:quality_objaverse}, while existing methods generate plausible appearances, they struggle with detail preservation. By directly optimizing appearance in 3D space, GAP achieves superior visual quality across object categories. A more detailed visual comparison with mesh-based methods is provided in the supplementary material.

\begin{table}[t!]
\vspace{-0.2cm}
   \caption{
    Quantitative comparison with baselines on the Objaverse dataset. Best results are highlighted as \colorbox{red}{1st}, \colorbox{orange}{2nd} and \colorbox{yellow}{3rd}.
   } 
   \vspace{-0.3cm}
   \label{tab:quantitative_objaverse}
   \small
   \centering
   \resizebox{\linewidth}{!}{
       \begin{tabular}{l|ccc|cc}
       \toprule
        \textbf{Method} & \textbf{FID$\downarrow$} & \textbf{KID$\downarrow$} & {\textbf{CLIP$\uparrow$}} & \multicolumn{2}{c}{\textbf{User Study} } \\
        & & & & \textbf{Overall Quality$\uparrow$} & \textbf{Text Fidelity$\uparrow$} \\ 
       \midrule
        \textbf{TexTure} \cite{richardson2023texture}  & 42.63 & 7.84 & \cellcolor{yellow} 26.84 & 2.90 & 3.05 \\ 
        \textbf{Text2Tex} \cite{chen2023text2tex} & 41.62 & 6.45 & 26.73 & \cellcolor{orange}3.48 & \cellcolor{orange}3.62 \\
        \textbf{SyncMVD} \cite{liu2024text}  & \cellcolor{orange}40.85 & \cellcolor{orange}5.77 & \cellcolor{orange}27.24 & \cellcolor{yellow}3.12 & \cellcolor{yellow}3.4 \\ 
        \textbf{Paint3D} \cite{zeng2024paint3d}  & \cellcolor{yellow}41.08 & \cellcolor{yellow}5.81 & 26.73 & 3.07 & 3.33\\ 
        \midrule
        \textbf{TexTure}\textsubscript{\textit{BPA}} & 60.69 & 15.98 & 26.62 & 1.46 & 1.62\\ 
        \textbf{Text2Tex}\textsubscript{\textit{BPA}} & 64.35 & 16.67 & 26.18 & 2.86 & 3.06 \\
        \textbf{SyncMVD}\textsubscript{\textit{BPA}}  & 60.29 & 14.35 & 26.19 & 2.85 & 3.12 \\ 
        \textbf{Paint3D}\textsubscript{\textit{BPA}}  & 65.36 & 17.37 & 25.14 & 1.45 & 1.45\\ 
        \midrule
        \textbf{TexTure}\textsubscript{\textit{CAP}}  & 53.55 & 12.43 & 26.68 & 2.23 & 2.60 \\ 
        \textbf{Text2Tex}\textsubscript{\textit{CAP}} & 52.78 & 11.09 & 26.78 & 3.03 & 3.57 \\
        \textbf{SyncMVD}\textsubscript{\textit{CAP}}  & 63.85 & 16.92 & 25.81 & 2.97 & 3.09 \\ 
        \textbf{Paint3D}\textsubscript{\textit{CAP}}  & 59.49 & 13.56 & 24.99 & 2.38 & 2.40 \\ 
        \midrule
        \textbf{Ours}  & \cellcolor{red}40.39 & \cellcolor{red}5.28 & \cellcolor{red}27.26& \cellcolor{red}4.21 & \cellcolor{red}4.47 \\  
        \bottomrule
       \end{tabular}
   }
       \vspace{-0.6cm}

\end{table}

\begin{figure*}[!h]
    \centering
    \includegraphics[width=\linewidth, alt={A detailed description of the method illustrated in the image}]{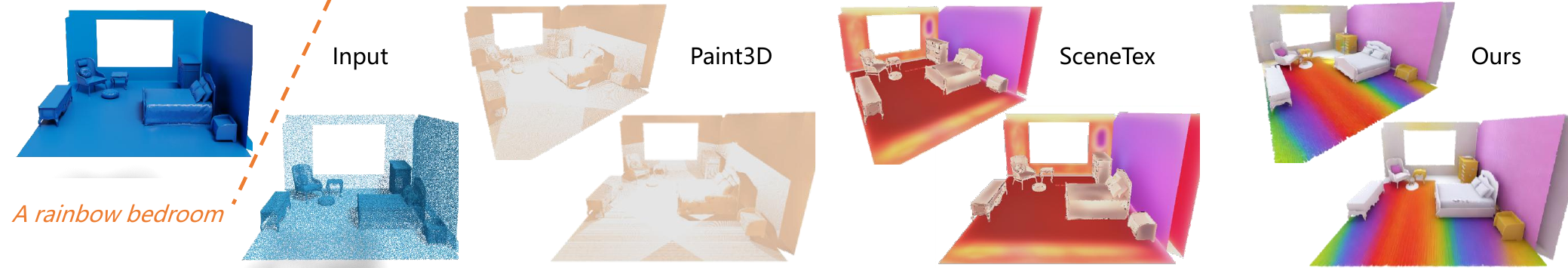}
    \vspace{-0.6cm}
    \caption{Scene-level Gaussianization comparison on 3D-FRONT datasets. }
    \vspace{-0.5cm}
    
    \label{fig:scene}
\end{figure*}



To assess visual appearance and text alignment, we conducted a user study with 30 participants. Each participant independently evaluated results from all methods across multiple viewpoints, rating them on a scale of 1 to 5.

\vspace{-0.2cm}
\subsection{Point-to-Gaussian Generation}
\label{sec.4.2}
\vspace{-0.2cm}
\noindent\textbf{Datasets and Implementations.} 
To evaluate GAP's effectiveness in Point-to-Gaussian generation, we conduct experiments on two datasets: the ShapeNet chair category \cite{chang2015shapenet} and DeepFashion3D \cite{zhu2020deep}. We uniformly sample 100K points from each 3D model to generate input point clouds. GAP is compared with three state-of-the-art methods DreamGaussian \cite{tang2023dreamgaussian}, TriplaneGaussian \cite{zou2023triplane}, and DiffGS \cite{zhou2024diffgs}, all using the same 100K point clouds as input. Please refer to the supplementary for the adaptions of those baseline methods, as well as additional results.

\noindent\textbf{Performance.} We provide visual comparisons with baseline methods in Fig.~\ref{fig:point2gaussian}, GAP consistently generates more visually appealing and geometrically accurate results compared to existing approaches. 
The baseline methods exhibit several key limitations. DreamGaussian, despite incorporating Score Distillation Sampling (SDS) for appearance optimization, tends to produce over-saturated appearances with unnatural colors. Additionally, its optimization process is computationally intensive and highly parameter-sensitive. TriplaneGaussian and DiffGS are fundamentally constrained by their limited-resolution triplane representations, limiting their ability to capture appearance details.


\vspace{-0.2cm}
\subsection{Gaussian Generation for Scanned Inputs}
\vspace{-0.1cm}
\label{sec.4.3}
\noindent\textbf{Datasets.} We evaluate GAP on real-world partial scans from SRB (Scan-to-Reality Benchmark) \cite{williams2019deep} and DeepFashion3D \cite{zhu2020deep} datasets. Both datasets contain point clouds captured by depth sensors, presenting real-world challenges such as incomplete coverage, occlusions and scanning artifacts. We directly use the raw scanned point clouds as input.

\begin{figure}[]
    \centering
    \includegraphics[width=0.9\linewidth, alt={A detailed description of the method illustrated in the image}]{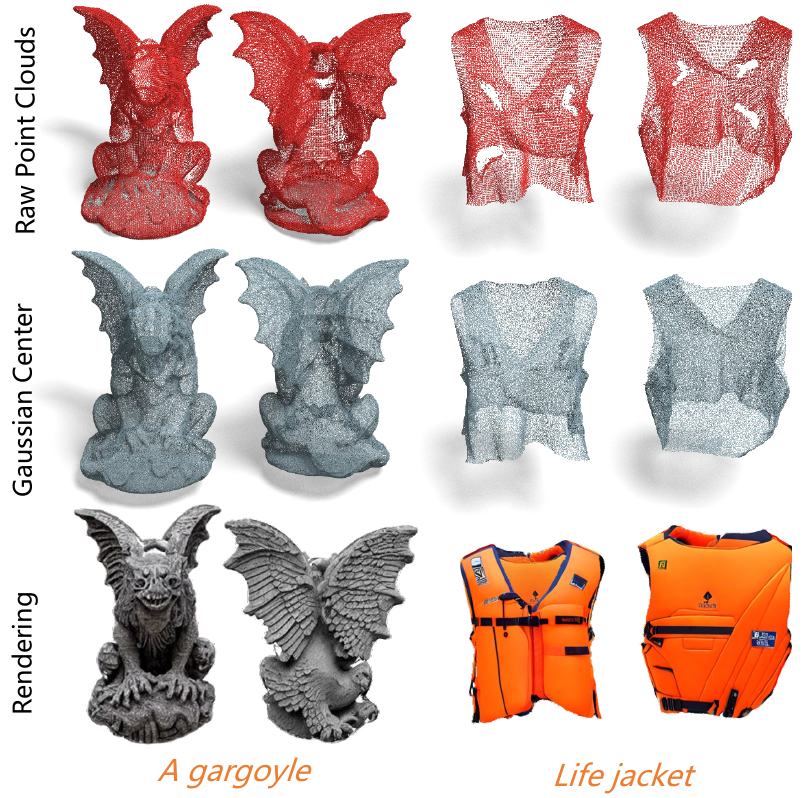}
    \vspace{-0.4cm}
    \caption{Results on real-world partial scans from SRB and DeepFashion3D datasets. }
    \label{fig:complementation}
    \vspace{-0.6cm}
    
\end{figure}

\noindent\textbf{Performance.} As shown in Fig.~\ref{fig:complementation}, GAP successfully gaussianizes partial point clouds into complete, high-quality Gaussian representations.  Our surface-anchoring mechanism effectively pull the split and cloned 3D Gaussians to fill missing regions while preserving geometric consistency. The results demonstrate that our method can robustly handle artifacts and occlusions in real-world scanned point clouds and generate visually appealing Gaussians.


\vspace{-0.1cm}
\subsection{Scale to Scene-Level Gaussian Generation} 
\vspace{-0.1cm}
\label{sec.4.4}
\noindent\textbf{Datasets.} We evaluate GAP on both synthetic and real-world scene datasets. For synthetic scenes, we use 3D-FRONT \cite{fu20213d}, which features diverse indoor environments. We sample 500K points from scene meshes as input. For real-world evaluation, we use raw point clouds from the 3D Scene dataset \cite{zhou2013dense}, which poses challenges such as complex topology, varying point densities, and scanning artifacts.

\noindent\textbf{Comparision.} Compared to Paint3D \cite{zeng2024paint3d} and Scenetex \cite{chen2024scenetex}, our method achieves superior visual quality. As shown in Fig.~\ref{fig:scene}, Paint3D fails on scene-level data, while SceneTex requires both VSD optimization \cite{wang2024prolificdreamer} and additional LoRA \cite{hu2021lora} training, significantly increasing processing time. In contrast, our method produces high-quality results for complex scenes with a single optimization process. Please refer to the supplementary for more results on real-world scenes.


\vspace{-0.1cm}
\subsection{Ablation Study}
\vspace{-0.1cm}
\label{sec.4.5}
\vspace{-2pt}
To analyze the effectiveness of key components in GAP, we performed a series of controlled experiments. The performance was measured using three metrics: FID, KID, and CLIP Score. These metrics were computed on rendered images captured from multiple viewpoints. We evaluate some major designs of our framework in Tab.~\ref{tab:ablation}. Without the Scale Loss, Gaussians grow excessively large, leading to distorted results in subsequent views. The Distance Loss prevents Gaussians from drifting away from object surfaces, maintaining geometric accuracy. The diffuse-based Gaussian Inpainting ensures complete coverage in hard-to-observe regions. Each component proves essential for achieving optimal performance.

\begin{table}[h]
   \vspace{-0.4cm}
   \caption{
    Ablation study of key components in GAP.
   } 
   \vspace{-0.3cm}
    \label{tab:ablation}
    \small
    \centering
    \setlength{\tabcolsep}{1.2em}
    \renewcommand{\arraystretch}{1.0}
    \begin{tabularx}{\linewidth}{>{\centering}m{3cm}| Y Y Y}
    \toprule
    \textbf{Method} & \textbf{FID$\downarrow$} & \textbf{KID$\downarrow$} & {\textbf{CLIP$\uparrow$}} \\
    \midrule
    \textbf{Full Model}  & \textbf{40.39} & \textbf{5.28} & \textbf{27.26} \\ 
    \midrule
    \textbf{W/o $\mathcal{L}_{\mathrm{Scale}}$}  &  214.63 & 79.04 & 26.25 \\ 
    \textbf{W/o $\mathcal{L}_{\mathrm{Distance}}$}  & 161.04 & 23.29 & 24.30 \\ 
    \textbf{W/o GS Inpainting}  & 46.37 & 8.77 & 27.21 \\ 
    \bottomrule
    \end{tabularx}
\vspace{-0.5\baselineskip}
\end{table}
\vspace{-0.5cm}
\section{Conclusion}
\vspace{-0.2cm}
In this paper, we presented GAP, a novel approach that generates high-quality 3D Gaussians from raw point clouds with text guidance. We design a multi-view optimization framework which learns Gaussian attributes from text-to-image diffusion models. The surface-anchoring constraint and diffuse-based Gaussian inpainting scheme are proposed to ensure geometric accuracy and appearance completion. Extensive experiments demonstrate GAP's effectiveness on both synthetic and real-world scanned data, from objects to large-scale scenes.

\clearpage

{
    \small
    \bibliographystyle{ieeenat_fullname}
    \bibliography{main}
}


\end{document}